\pdfoutput=1

\documentclass[11pt]{article}

\usepackage{emnlp2021}
\usepackage{multirow}
\usepackage{times}
\usepackage{latexsym}
\usepackage{amsmath}
\usepackage{graphicx}
\usepackage{mathtools, nccmath}
\usepackage[T1]{fontenc}

\usepackage[utf8]{inputenc}

\usepackage{microtype}

\title{SWEAT: Scoring Polarization of Topics across Different Corpora}

\author{Federico Bianchi \\
  Bocconi University \\
  Via Sarfatti 25 \\
  Milan, Italy \\
  \texttt{f.bianchi@unibocconi.it} \\\And
    Marco Marelli \\
  University of Milano-Bicocca \\
  Piazza dell’Ateneo Nuovo, 1 \\
  Milan, Italy \\
  \texttt{marco.marelli@unimib.it} \\\AND
  Paolo Nicoli \\
  University of Milano-Bicocca \\
  Viale Sarca 336 \\
  Milan, Italy \\
  \texttt{p.nicoli@campus.unimib.it} \\ \And
  Matteo Palmonari \\
  University of Milano-Bicocca \\
  Viale Sarca 336 \\
  Milan, Italy \\
  \texttt{matteo.palmonari@unimib.it} \\ 
  }

\begin{document}
\maketitle
\begin{abstract}
Understanding differences of viewpoints across corpora is a fundamental task for computational social sciences. In this paper, we propose the Sliced Word Embedding Association Test (SWEAT), a novel statistical measure to compute the relative polarization of a topical wordset across two distributional representations. To this end, SWEAT uses two additional wordsets, deemed to have opposite valence, to represent two different poles. We validate our approach and illustrate a case study to show the usefulness of the introduced measure.
\end{abstract}

\section{Introduction}

In this short paper, we introduce a method to score polarization of different corpora with respect to a given topic. The method is intended to support studies where two different corpora are compared (e.g., news sources inspired by different political positions or social communities characterized by different viewpoints) to investigate whether they convey implicit attitudes towards a given topic. This corpus-wise comparison - its main peculiarity with respect to the large body of work proposed to study and correct bias in NLP models (we refer to~\cite{garrido2021survey} for a detailed survey on bias in NLP models) - is based on a new measure that we introduce, the Sliced Word Embedding Association Test (SWEAT).          

SWEAT is an extension of the Word Embedding Association Test (WEAT) proposed by \newcite{caliskan2017semantics}, which measures the comparative polarization for a pair of topical wordsets (e.g., insects and flowers) against a pair of attribute wordsets (e.g., pleasant and unpleasant) in a single-corpus distributional model (e.g. 1950 American newspaper articles).  In this context, with polarization we refer to the phenomenon for which two communities have opposite attitudes against some topic.

With SWEAT we extend this approach by measuring the relative polarization for a single topical wordset - \textit{the topic}, using a pair of stable attribute wordsets deemed to have opposite valence - \textit{the two poles}, in a pair of aligned distributional models representing the semantics of two different corpora. We explain the rationale behind SWEAT with an example. Suppose that we want to investigate whether two different Italian news sources, e.g., La Repubblica (known to be closer to center-left political positions) and Il Giornale (known to be closer to center-right political positions) hold different and opposite viewpoints about a topic, e.g., ``Berlusconi'' (a reference center-right Italian politician in the recent past). We can collect a news corpus from La Repubblica and one from Il Giornale to train two different distributional models in such a way that they are aligned~\cite{hamilton-etal-2016-diachronic,di2019training,cassani2021word}. 

We expect that some words have stable meanings while other change across corpora reflecting the different viewpoints. We can then select a set of words describing the ``Berlusconi'' topic, whose representations are expected to differ across corpora, and two wordsets having respectively positive and negative valence (the two poles), whose representations are expected to be stable across corpora. The main idea behind SWEAT is the following: if the two corpora hold polarized views about the topic, the ``Berlusconi'' wordset will be associated more strongly with the positive rather than with the negative pole in one corpus (Il Giornale), while the opposite association will hold in the other corpora (La Repubblica). SWEAT measures this difference and reports effect size and significance.                   

\paragraph{Contributions.} We introduce the SWEAT, a novel statistical measure to study relative polarization in distributional representations. We additionally introduce a lexicon selection pipeline and an easy-to-use code to create visualizations. We believe our measure can be useful for different use cases in the computational social science field.  We share a repository with an easy to use implementation of our measure.\footnote{\url{https://github.com/vinid/SWEAT}}

\section{Background: WEAT}

\newcite{caliskan2017semantics} introduce the Word Embedding Association Test (WEAT) to test whether distributional representations exhibited the same implicit biases detected in social sciences studies through behaviorally-measured word associations~\cite{greenwald1998measuring}.

The WEAT compares the relative associations of a \textit{pair} of target concepts $X$ and $Y$ (e.g., Science and Arts) to a \textit{pair} of attribute concepts $A$ and $B$ (e.g., Male and Female) in a distributional vector space $\mathcal{E}$;  $X$, $Y$, $A$, and $B$ are all sets that contain representative words for the concept. The statistical measure is based on the following formula:

\begin{equation}
\small
S(X, Y, A, B) = \sum_{x \in X} s(x, A,B) - \sum_{y \in Y}s(y, A, B)  \nonumber
\end{equation}

The value $s(w, A, B)$ is instead computed as:

\begin{equation}
\small
\frac{1}{|A|}\sum_{a \in A} \cos(\mathcal{E}(w), \mathcal{E}(a)) - \frac{1}{|B|}\sum_{b \in B} \cos(\mathcal{E}(w), \mathcal{E}(b))  \nonumber
\end{equation}

Where the effect size is defined as: 
\begin{equation}
d = \frac{\text{mean}_{x \in X} s(x,A,B) - 
\text{mean}_{y \in X} s(y,A,B)}{\text{std}_{w \in X \cap Y}s(w,A,B)} \nonumber
\end{equation}

Significance is computed through a permutation test~\cite{dwass1957modified} over the possible partition of equal size for the union of target-wordsets $\mathcal{P}\left[ X \cup Y \right] = \{ (X_i,Y_i) \}_i$. The $p$-value is computed as the rate of scores, from all possible permutations, that are higher than the tested one: 
$P_i[S(X_i,Y_i,A,B) > S(X,Y,A,B) ]$.

Depending on the sign of the score the association could be either $X\sim A, \: Y\sim B$ for positive scores and $X \sim B, \: Y \sim A$ for negative ones. where the $\sim$ indicates semantic association. 

\section{SWEAT Pipeline}

Our SWEAT pipeline is composed of two main components: a statistical test to evaluate relative polarization and a lexicon refinement pipeline - based on aligned word embeddings - to help researchers select the pole wordsets.


\subsection{Measuring Polarization with SWEAT}
\label{subs:SWEAT_def}
The SWEAT measure operates following the same structure as the WEAT with one key difference: given two corpora SWEAT uses two corpus-specific distributional representations $\mathcal{E}^1$ and $\mathcal{E}^2$ instead of one. Thus, the relative mean associations depend explicitly on the corpus-specific embedding functions that map words to the respective embedding spaces.


We define the SWEAT score $S(W,\mathcal{E}^1,\mathcal{E}^2,A,B)$ as follows: 

\begin{equation}
\sum_{w\in W} s(w,\mathcal{E}^1,A,B)  - \sum_{w\in W} s(w,\mathcal{E}^2,A,B) \nonumber \\ \end{equation}

$s(w,\mathcal{E},A,B)$ is computed as

\begin{equation}
\begin{split}
    s(w,\mathcal{E},A,B) = \frac{1}{\vert A \vert }\sum_{a\in A} \cos \left( \mathcal{E}(w),\mathcal{E}(a)\right) \\ - \frac{1}{\vert B \vert }\sum_{b\in B} \cos \left( \mathcal{E}(w),\mathcal{E}(b)\right) \nonumber
\end{split}
\end{equation}

where $W$ is the topical-wordset and $A$ and $B$ are the pole-wordsets. Similarly as in WEAT, the score sign indicates the detected associations: a positive sign indicates that the representations in $\mathcal{E}^1$ of the target wordset are relatively more associated with the pole-wordset $A$ than the representations of the same words in $\mathcal{E}^2$, while a negative sign indicates the opposite. The effect size and significance level follow the same structure as the WEAT and are omitted for brevity. 

Observe that the wordsets A and B are not forced to represent opposite concepts (e.g., positive and negative), but, if they are, SWEAT provides a measure for scoring polarization.

To preserve comparability between the embedding spaces, it is important to align them, a task that is common in temporal word embeddings~\cite{hamilton-etal-2016-diachronic,di2019training} and in cross-lingual word embeddings~\cite{ruder2019survey}. 
The alignment step ensures the comparability of the slices when computing the similarities and supports lexicon refinement as described below.

\subsection{Lexicon Refinement}\label{subs:lrp}
The definition of the pole-wordsets ($A$ and $B$) can be supported by the use of lexica. However, words in a lexicon might have different meanings depending on the context in a representation $\mathcal{E}^k$; Thus, we first apply a refinement method based on the aligned word embedding spaces. We leverage aligned word embeddings to filter out unstable words from the lexicon. We keep a word in the lexicon if moving its word vector from $\mathcal{E}^1$ to $\mathcal{E}^2$ brings us to the same word and vice-versa. This process is done to ensure that the selected pole words do not change the representation between corpora. 

Moreover, low-frequency words tend to produce lower-quality representations. We apply the Zipf~\cite{van2015call} relative frequency measure to exclude such low-frequency words from the pole-wordsets; following the recommendations in~\cite{van2015call}, we select words with Zipf score > $5$.







\subsection{Visualization}
SWEAT comes with easy to create visualizations. For example Figure~\ref{fig:SWEAT_cumulative_experiment} shows four cumulative visualizations illustrating four explored relations. For each model, the two color-coded areas indicate the sums, over all the topical-wordset elements $x\in X$, of the mean associations to the two the pole-wordsets $A$ and $B$, one for each color.

The black \textit{cumulate} dot over each bar-chart indicates the sum of its two color-coded parts.

These visualizations should not be used to directly compare different SWEAT scores because the horizontal scale adapts to the effects' magnitude; effect sizes can be used instead.

\subsection{Interpretation of SWEAT}

To conclude, we can summarize how to interpret SWEAT scores and plots as follows. 
SWEAT measures a difference in the association between corpus-specific representations of a topical-wordset and two attribute wordsets. When the effect size is high in absolute value, the difference is also high. When attribute wordsets represents two poles, a high difference in association reveals different attitudes towards the two pole-wordsets. When the cumulative associations between the corpus-specific representations of the topic and either pole-wordset have different signs, SWEAT reveals a polarization effect in addition to a plain difference.

\section{Validation on the Reddit Corpus}

We validated the SWEAT measures using subcorpora from the Reddit Corpus.\footnote{\texttt{\url{https://archive.org/details/2015_reddit_comments_corpus}}} The corpus, previously used by \newcite{hamilton-etal-2016-inducing}, contains all the posts on the social media platform Reddit for the year 2014.

We selected for this experiment the English boards \textit{AnarchoCapitalism}-\textit{Frugal} and \textit{BabyBumps}-\textit{Childfree}.
The first pair is centered around topics of finance and capitalism, with AnarchoCapitalism (\textbf{AC}) expected to have positive polarization, due to its pro-capitalism beliefs, and Frugal (\textbf{FR}) negative.

The second pair is focused on parenting (i.e. having children), with BabyBumps (\textbf{BB}) expressing positive opinions about this matter and ChildFree (\textbf{CF}) being negatively polarized. We do pairwise comparisons using the SWEAT measure. The embeddings are generated and aligned using TWEC~\cite{di2019training,cassani2021word}.

\paragraph{Lexicon and Wordset Selection.}

For the pole-wordsets, we used the Inquirer Sentiment Lexicon~\cite{stone1966general} that was selected for its large size and generality. We used the pipeline described in the above section as a filter.

The topical wordsets for each pair, AC-FR and BB-CF, were selected in a semiautomatic fashion (see Appendix) in order to manually remove ambiguous terms which could bias the final result (e.g., the term \textit{fine} is polysemous denoting both a \textit{financial penalty} and \textit{good}, and thus had to be removed from both pole and topical wordsets to prevent undue associations). The two derived topic wordset for \textit{capitalism} and \textit{parenting} are presented below:

\begin{figure*}
\centering
\includegraphics[width=0.85\textwidth]{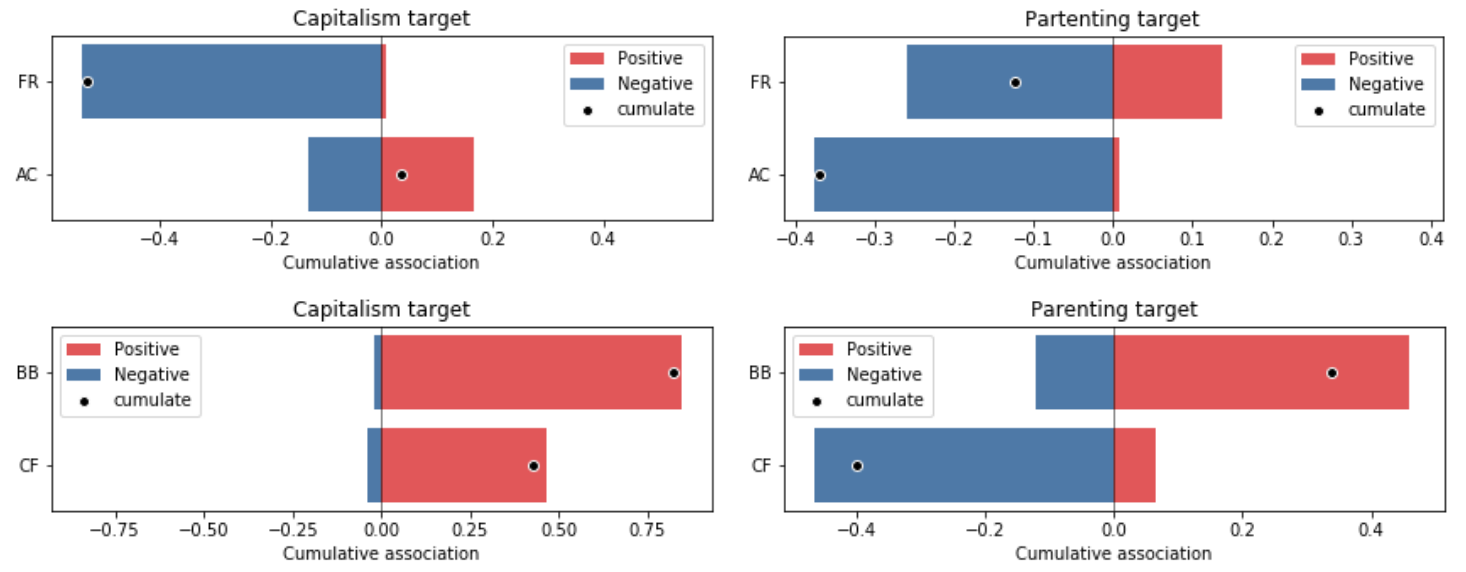}
\caption{Cumulative SWEAT visualizations for experimental results.}
\label{fig:SWEAT_cumulative_experiment}
\end{figure*}

\begin{quote}
\textbf{capitalism}: \textit{\textit{market}, \textit{value}, \textit{economic}, \textit{capital}, \textit{price}, \textit{wealth}, \textit{profit}, \textit{companies}, \textit{interest}, \textit{cost}, \textit{competition}, \textit{trade}}

\textbf{parenting}: \textit{\textit{baby}, \textit{birth}, \textit{child}, \textit{daughter}, \textit{family}, \textit{father}, \textit{kid}, \textit{mother}, \textit{pregnant}, \textit{son}, \textit{parent}, \textit{children}}
\end{quote}


\paragraph{Results and Visualizations}
Table~\ref{tab:SWEAT_results} shows the results for each subreddit pair. Statistically significant relative polarizations are detected on the appropriate topics only ($\alpha=0.01$): polarization in the parenting topic is observed when considering the ChildFree-vs-BabyBumps subreddits and not when considering the AnarchoCapitalis-vs-Frugal subreddits, and vice-versa when considering the financial topic. Effect sizes are also coherent with the experiment setup, 
presenting very large~\cite{cohen1988statistical} magnitudes for the two relations of interest.

\begin{table*}[ht]
\centering
\small
 \begin{tabular}{| c | c | c | c | c | c |}
 \hline
 \textbf{Corpora} & \textbf{Topic} & \textbf{SWEAT} & \textbf{eff. size} & \textbf{p-value} & \textbf{associations} \\
 \hline
 \multirow{4}{*}{\textit{AC},\textit{FR}} & \multirow{2}{*}{\textbf{capitalism}} & \multirow{2}{*}{0.5661} & \multirow{2}{*}{1.3123} & \multirow{2}{*}{\textbf{0.0052}} & $AC\sim\oplus$ \\
 \multirow{4}{*}{}  & \multirow{2}{*}{} & \multirow{2}{*}{} & \multirow{2}{*}{} & \multirow{2}{*}{} & $FR\sim\ominus$ \\
 \cline{2-6}
 \multirow{4}{*}{}  & \multirow{2}{*}{parenting} & \multirow{2}{*}{-0.2457} & \multirow{2}{*}{-0.6443} &  \multirow{2}{*}{0.1479} & $AC\sim\ominus$\\
 \multirow{4}{*}{}  & \multirow{2}{*}{} & \multirow{2}{*}{} & \multirow{2}{*}{} & \multirow{2}{*}{} & $FR\sim\oplus$ \\
 \hline
 \multirow{4}{*}{\textit{CF}, \textit{BB}} & \multirow{2}{*}{capitalism} & \multirow{2}{*}{-0.3982} & \multirow{2}{*}{-0.6936} &  \multirow{2}{*}{0.1196} & $CF\sim\ominus$ \\
 \multirow{4}{*}{}  & \multirow{2}{*}{} & \multirow{2}{*}{} & \multirow{2}{*}{} & \multirow{2}{*}{} & $BB\sim\oplus$ \\
 \cline{2-6}
 \multirow{4}{*}{}  & \multirow{2}{*}{\textbf{parenting}} & \multirow{2}{*}{-0.7391} & \multirow{2}{*}{-1.2891} & \multirow{2}{*}{\textbf{0.0079}} & $CF\sim\ominus$\\
 \multirow{4}{*}{}  & \multirow{2}{*}{} & \multirow{2}{*}{} & \multirow{2}{*}{} & \multirow{2}{*}{} & $BB\sim\oplus$ \\
 \hline
 \end{tabular}
\caption{SWEAT experiments results ($\oplus$ and $\ominus$ indicate positive and negative polarization respectively).}
\label{tab:SWEAT_results}
\end{table*}

Figure~\ref{fig:SWEAT_cumulative_experiment} presents the four cumulative visualizations illustrating the four explored relations. The Figure shows two centered horizontal stacked bar-charts, one for each aligned model $\mathcal{E}^1$, $\mathcal{E}^2$. 
The two significant polarizations can be seen on the main diagonal: in the top left for the 	\textit{capitalism} topic the \textbf{FR} corpus can be seen being strongly negative while the \textbf{AC} one has an overall positive but more nuanced position; in the bottom right, the polarization difference over the \textit{parenting} topic is much more pronounced, with \textbf{CF} carrying a strong negative polarization in stark contrast with the positive one expressed by \textbf{BB}, thus confirming our hypotheses of polarization of the subreddits.

\section{Case Study on Italian Newspapers}
We provide an analysis on \textit{Silvio Berlusconi} that focuses on the homonymous politician who, among other things, is the indirect owner of the Italian \textit{Il Giornale} (GIO) newspaper and is, as an exponent of the center-right coalition, seldom praised by more left-leaning newspaper such as \textit{La Repubblica} (REP). The dataset (see Appendix) contains a collection of roughly 40 thousand articles in Italian from March 2018 to June 2019 for REP and GIO.

As a lexicon we used \textit{Sentix}~ \cite{BasileNissim2013WASSA}. Only the elements meeting all the following empirical criteria were considered in the final lexicon: POS tagging as either \textit{adjective} or \textit{verb}; \textit{intensity} equal or greater to 0.75; either \textit{positive} or \textit{negative} score equal or greater to 0.75; \textit{polarity} equal to $\pm1$ and that are single-word lemma. We then applied the lexicon filtering pipeline.



We conducted the analysis on the topic \textit{Silvio Berlusconi} manually selecting the following topical-wordset:

\begin{quote}
\textbf{Berlusconi}: \textit{\textit{cavaliere}, \textit{berlusconi}, \textit{arcore}, \textit{mediaset}, \textit{fininvest}, \textit{silvio}, \textit{rete4}, \textit{fi}, \textit{pdl}, \textit{iene}, \textit{vespa}, \textit{tg5}} 
\end{quote}

\begin{figure}
\centering
\includegraphics[width=1\columnwidth]{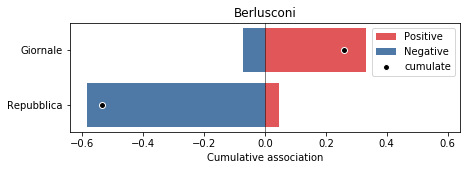}
\caption{Cumulative SWEAT visualizations for the Berlusconi wordset.}
\label{fig:spdati_SWEAT_berlusconi}
\end{figure}

The polarization detected by the SWEAT analysis confirms a significantly more positive polarization in the GIO corpus and a more negative one in REP (SWEAT $=-0.795$, $d=-1.3276$, \textit{p-value} $=0.0038$). 
Figure~\ref{fig:spdati_SWEAT_berlusconi} shows the results of this association and confirms our general intuition.

\section{Limitations}\label{sec:limitations}
The SWEAT is able to capture relative semantic polarizations of topic wordsets. However, the measure is not suitable for fully unsupervised data-driven inference on the polarization of a corpora collection, similarly to the WEAT measure. 
The measure results are dependent on the choice of topic wordsets, and it is possible to find that some wordset is significantly polarized even though its elements are not semantically related to each other. For this reason, it is advised that researchers exercise caution in selecting which words to include in the topic wordset, and which to exclude.

\section{Related Work}
There has been great attention to the problem of polarization and bias in word embeddings in the latest years. 
We refer the reader to~\cite{garrido2021survey,blodgett2020language} for relevant works. 

SWEAT and the applications we have shown have also a strong connection with the body of work related to the detection of hyper-partisan media~\cite{kiesel-etal-2019-semeval,bestgen-2019-tintin}, where extreme left-wing or right-wing positions of the news have to be detected.

We want to mention that the introduction of the WEAT measure has have been of great influence on the community. Indeed recently, \newcite{zhou-etal-2019-examining} extended WEAT to have specific support for gendered languages (where different terms can have male and female counterparts). \newcite{chaloner-maldonado-2019-measuring} use the WEAT measure to show how different domains contain different biases. Instead, \newcite{lauscher-glavas-2019-consistently} propose XWEAT as a cross-lingual version of the WEAT measure. 

\section{Conclusions}

We have described a new measure to compute relative polarization on corpus-specific distributional representations. Measuring implicit attitudes is a crucial endeavor in social sciences, and in particular social psychology~\cite{de2009implicit}. Recent developments have shown that such attitudes are not only observable in human behavior, but can also be captured via text analyses~\cite{caliskan2017semantics,bhatia2017semantic}. SWEAT brings a further advancement in this research line by allowing to validate hypotheses on the polarization of content from different text sources.

\section*{Ethical Considerations}
SWEAT can be used to capture relative polarization and we are aware that the measures can also be used to test the significance of ethical polarization in different contexts. As we remark in Section~\ref{sec:limitations} the measure we implemented comes with some limitations that have to be considered during experiments.

\section*{Acknowledgments}
Federico Bianchi is a member of the Bocconi Institute for Data Science and Analytics (BIDSA) and the Data and Marketing Insights (DMI) unit. We would also like to thank SpazioDati\footnote{\url{https://spaziodati.eu}} for providing the Italian newspaper corpora. The research at University of Milano-Bicocca has been supported in part by EU H2020 projects EW-Shopp - Grant n. 732590, and EuBusiness-Graph - Grant n. 732003

\bibliography{anthology,custom}
\bibliographystyle{acl_natbib}
\appendix

\section{Replication: Data Details}

\subsection{Reddit}
The Reddit corpus used is a publicly available\footnote{\url{https://archive.org/details/2015_reddit_comments_corpus}} collection of posts published on the social media platform Reddit between 2007 and 2015.

As a first pre-processing step the entirety of the files were parsed and grouped by subreddit. During this step posts containing the \texttt{[removed]} or \texttt{[deleted]} keywords (indicating that the post was either removed by the moderation team or deleted by the user) were omitted from processing; additionally a list of known bot accounts was used to prune their posts from the analysis.

Having divided the posts into subreddit corpora the top 250 by size were selected, following the same procedure described by~\newcite{hamilton-etal-2016-inducing}. The resulting corpora were further pre-processed though case-folding and removal of punctuation and tags using the \texttt{gensim}\footnote{\url{https://radimrehurek.com/gensim/}} python text processing library.

\paragraph{Topical Candidate Selection} An initial set of candidate words was identified by collecting the top-100 most frequent\footnote{ordered by average Zipf measure} words from the shared vocabulary of all four corpora after removing stopwords and language operators (i.e. elements with little semantic connotation like auxiliary verbs or adverbs).

The topical wordset was then compared with the polarization wordsets for that pair, removing ambiguous terms, and lastly a manual separation into topics was performed to identify the true wordset of interest and removing other subtopics (for example, the candiate wordset for \textit{parenting} also included terms related hospitals and medicine or the \textit{capitalism} one had terms related to cryptocurrencies).

\subsection{Italian NewsPapers}

Data from Italian Newspapers have been kindly provided by \textit{SpazioDati}.\footnote{\url{https://spaziodati.eu/en/}} We removed punctuation from the text and made it lowercase.

We used the following empirical criteria to select a subset of the Sentix lexicon: 
POS tagging as either \textit{adjective} or \textit{verb}; \textit{intensity} equal or greater to 0.75; either \textit{positive} or \textit{negative} score equal or greater to 0.75; \textit{polarity} equal to $\pm1$ and where single-word lemma. This is the list of words we extracted: 

\begin{quote}
\textbf{positive}: \textit{\textit{meglio}, \textit{bello}, \textit{migliore}, \textit{considerato}, \textit{giusto}, \textit{felice}, \textit{importante}, \textit{grande}, \textit{semplice}, \textit{maggiore}}

\textbf{negative}: \textit{\textit{morto}, \textit{peggio}, \textit{difficile}, \textit{vecchio}, \textit{impossibile}, \textit{pericoloso}, \textit{male}, \textit{purtroppo}}
\end{quote}

\section{Replication: Experiment Parameters}
To generate the aligned word embedding representations, we train TWEC~\cite{di2019training} using a dimension of the embeddings equal to 100 and a window size of 5. Iterations are set to 5 for both the static and the dynamic iterations.

\section{Details: Visualization}\label{subs:SWEAT_viz}

\begin{figure*}[h]
\centering
\includegraphics[width=0.5\textwidth]{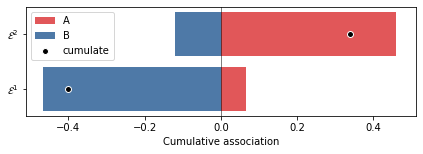}
\caption{Cumulative SWEAT visualization for a case of strong polarization $\mathcal{E}^1\sim B, \:\mathcal{E}^2 \sim A$.
}
\label{fig:SWEAT_cumplot}
\end{figure*}

\begin{figure*}[h]
\centering
\includegraphics[width=0.8\textwidth]{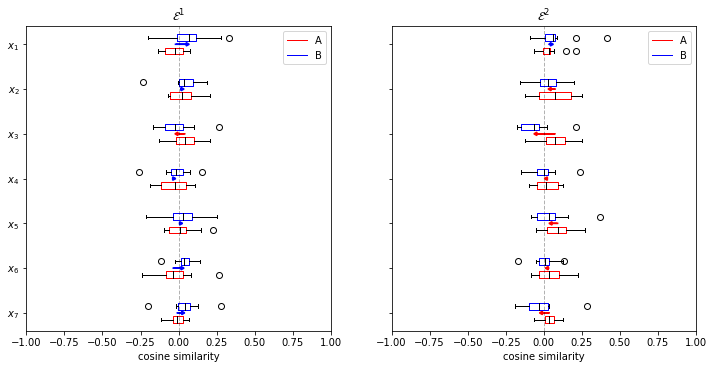}
\caption{Exploded views of SWEAT association distributions between topical words and pole wordsets}
\label{fig:SWEAT_detail}
\end{figure*}

To aid the exploration and interpretation of SWEAT results the framework implements two main visualizations: an aggregate plot for the two models of the cumulative associations to the two pole-wordsets (Figure~\ref{fig:SWEAT_cumplot}); a detailed view of the association distribution over the two pole-wordsets for each element of the topical-wordset (Figure~\ref{fig:SWEAT_detail}).

The first visualization (Figure~\ref{fig:SWEAT_cumplot}) shows two centered horizontal stacked bar-charts, one for each aligned model $\mathcal{E}^1$, $\mathcal{E}^2$. 
For each model the two color-coded areas $\beta_A, \beta_B$ indicate the sums, over all the topical-wordset elements $w\in W$, of the mean associations to the two the pole-wordsets $A$ and $B$, one for each color. 

The black \textit{cumulate} dot over each bar-chart indicates the sum of its two color-coded parts, i.e. the $\sum_w s(w,\mathcal{E}^k,A,B)$ relative to the model $\mathcal{E}^k$ associated to that bar-chart.
The final score is not directly encoded in the visualization, but is given by the difference between the position of the first and second cumulate dots.

The first visualization can also be broken down in its main parts, for which we provide a second visualization (Figure~\ref{fig:SWEAT_detail}) also divides the two aligned models $\mathcal{E}^1$, $\mathcal{E}^2$ into distinct views, this time two side by side canvasses of pairs of color-coded boxplots\footnote{In a slight deviation from the usual boxplot design, here the box \textit{belt} indicates the \textit{mean} instead of the \textit{median} to better illustrate the SWEAT measure inner formula elements}. 
In each canvass a pairs of boxplots represent, for an element of the topical-wordset $x_i \in X$, the two distributions $\delta_A^{x_i}, \delta_B^{x_i}$ of its associations to the pole-wordsets $A$ and $B$, using the same color-coding as the first visualization; across the two canvasses pairs are horizontally aligned, indicating the association distributions of the same topical word in the two models.
\begin{equation}
\delta_P^{x_i} = \left\lbrace \cos(x_i,p) \right\rbrace_{p \in P} \; P \in \{A,B\}
\end{equation}

Additionally for each pair of boxplots a color-coded arrow connects the two means to help  illustrate the relative mean association $s(x_i,\mathcal{E}^k,A,B)$ for that word: the arrow always connects $A\rightarrow B$ and takes the color of the \textit{dominant} pole, i.e. the one with the stronger mean association which in turn contributes in the cumulative sum for the final score.

\section{Details: Computing Infrastructure}
We ran the experiments on a common laptop. All the computations are run on CPU, the model name is: Intel(R) Core(TM) i7-8750H CPU @ 2.20GHz. The running time for the SWEAT measure is bound by the number of words in the wordset and in the lexicon; however the computation of the SWEAT described in this paper take few minutes.

\end{document}